\documentclass[sn-mathphys-num]{sn-jnl}

\usepackage{graphicx}%
\usepackage{multirow}%
\usepackage{amsmath,amssymb,amsfonts}%
\usepackage{amsthm}%
\usepackage{mathrsfs}%
\usepackage[title]{appendix}%
\usepackage{xcolor}%
\usepackage{textcomp}%
\usepackage{manyfoot}%
\usepackage{booktabs}%

\usepackage{listings}%

\theoremstyle{thmstyleone}%
\theoremstyle{thmstyletwo}%

\theoremstyle{thmstylethree}%

\begin{document}

\title[Lehmer Neural Networks]{Efficient and Interpretable  Neural Networks\\ Using Complex Lehmer Transform}


\author*[1]{\fnm{Masoud} \sur{Ataei}}\email{masoud.ataei@utoronto.ca}

\author[2]{\fnm{Xiaogang} \sur{Wang}}\email{stevenw@yorku.ca}


\affil*[1]{\normalsize\orgdiv{Department of Mathematical and Computational Sciences},\\ \orgname{University of Toronto}, \state{Ontario}, \country{Canada}}

\affil[2]{\normalsize\orgdiv{Department of Mathematics and Statistics}, \orgname{York University}, \state{Ontario}, \country{Canada}}



\abstract{We propose an efficient and interpretable neural network with a novel activation function called the weighted Lehmer transform. This new activation function enables adaptive feature selection and extends to the complex domain, capturing phase-sensitive and hierarchical relationships within data. Notably, it provides greater interpretability and transparency compared to existing machine learning models, facilitating a deeper understanding of its functionality and decision-making processes.  We analyze the mathematical properties of both real-valued and complex-valued Lehmer activation units and demonstrate their applications in modeling nonlinear interactions. Empirical evaluations demonstrate that our proposed neural network achieves competitive accuracy on benchmark datasets with significantly improved computational efficiency. A single layer of real-valued or complex-valued Lehmer activation units is shown to deliver state-of-the-art performance, balancing efficiency with interpretability. }




\maketitle

\newpage
\section{Introduction}
\label{sec1} 
We might have already entered the era of superintelligence, in which machines achieve better cognitive capabilities than humans. The current machine learning methods have made many revolutionary advances that were not even imaginable just a few years ago. The current deep learning models, while effective in capturing intricate feature interactions, often rely on numerous layers to achieve high performance with millions or even billions of parameters to capture complicated patterns. The black-box nature of the current machine learning architect has always been prohibitive for human reasoning to truly understand the nature of those millions of operations. This results in increased computational cost, diminished transparency, and challenges in understanding the underlying mechanisms driving their predictions. More importantly, this brings significant unforeseeable risks and uncertainties if we blindly trust the outcomes of those large models. For human intelligence to remain relevant in building machine learning models and to greatly reduce computational costs, we would need to construct neural network architectures that are not only efficient but also interpretable.

Inspired by the Kolmogorov-Arnold representation theorem, a Kolmogorov-Arnold Network (KAN) has been previously proposed with a learnable node activation function \cite{liu2024kan}. This network serves as an alternative to the traditional multi-layer perceptron architecture by leveraging the theorem's elegant decomposition of high-dimensional functions. However, learning a high-dimensional function through a one-dimensional function can be challenging due to potential non-smoothness or fractal-like behavior in some mappings, which complicates optimization and interpretability. The KAN mitigates these issues by introducing a learnable activation function that adapts to the complexity of the target mappings. In this work, we take a different approach by employing a fixed nonlinear activation function within a single-layer architecture. This design not only simplifies the optimization process but also enhances interpretability and computational efficiency, offering a compelling alternative to KAN for applications requiring clarity and scalability.

Specifically, we propose an efficient and interpretable neural network using a novel activation function called the Lehmer transform. The Lehmer transform, a parameterized family of means, provides a robust framework for summarizing and transforming features \cite{ataei2022theory}. Its ability to interpolate between different forms of mean behaviors allows it to emphasize small or large elements dynamically, making it a versatile tool for data representation. Building on its foundational mathematical properties, we extend the Lehmer transform into a weighted form, enabling adaptive feature importance through trainable weights. Additionally, we generalize the transform to the complex domain, introducing an oscillatory component that captures phase-sensitive dependencies and hierarchical relationships within data. The weighted form reflects the relative connection among inputs and also their absolute scale. It also enjoys perturbation and proportion invariance as well as smoothness and Schur convexity. The complex-valued weighted Lehmer transform introduces great flexibility by introducing an oscillatory component and enjoys differentiability in the complex domain. It also has a built-in smoothing or summarizing functionality for the inputs.

These extensions form the foundation of the Lehmer Neural Network (LNN), a novel architecture that leverages the unique properties of the Lehmer transform. Central to the LNN are the Lehmer Activation Units (LAUs), which enable nonlinear summarization and phase-sensitive transformations through real-valued and complex-valued variants. The real-valued LAUs provide a parameterized mechanism for hierarchical feature aggregation, while the complex-valued LAUs further incorporate oscillatory contributions, enhancing their representational power for datasets with structured or phase-sensitive characteristics. With one single layer, this novel Neural Network achieves competitive accuracy with classical models on structured datasets such as Iris, Wine, and Wisconsin Breast Cancer, as well as high-dimensional datasets like MNIST. Empirical evaluations on these benchmark datasets also demonstrate the efficacy of LAUs in diverse contexts. These results highlight the ability of the LNN to achieve high performance with reduced architectural complexity, showcasing its efficiency and adaptability.

To contextualize the development of our proposed neural network, it is helpful to examine recent advancements in interpretable neural networks. The increasing complexity and opacity of modern machine learning models have intensified the focus on interpretability, particularly in high-stakes domains such as healthcare and finance. Techniques such as attention mechanisms \cite{bahdanau2015neural} and saliency maps \cite{simonyan2014deep} have become prominent tools for providing insights into feature importance and decision-making processes. Recent advancements in self-explaining neural architectures embed interpretability directly into the network \cite{alvarez2018towards}, ensuring transparency without compromising predictive accuracy. Concept-based methods \cite{kim2018tcav} and graph neural networks (GNNs) \cite{kipf2017semi} further extend these approaches, enabling interpretable learning in complex and structured data. These methods address the growing need for models that are both effective and explainable, particularly for ethical and transparent AI deployment \cite{rudin2019stop, doshi2017towards}.

The opacity of traditional neural networks has also motivated the development of post-hoc explainability methods such as Layer-wise Relevance Propagation (LRP) \cite{montavon2017explaining} and Local Interpretable Model-agnostic Explanations (LIME) \cite{ribeiro2016should}. These techniques analyze feature importance after training, providing intuitive explanations for model behavior. Despite their widespread adoption, these approaches often fail to address the underlying training dynamics or ensure interpretability during the decision-making process. Recent research highlights the importance of integrating interpretability directly into the model architecture, exemplified by attention-based models \cite{vaswani2017attention} and hybrid frameworks that combine interpretable components with high-performance black-box systems \cite{gilpin2018explaining}. This integration offers a promising path forward, balancing the trade-off between interpretability and predictive power \cite{roscher2020explainable, tjoa2020survey}.

Simultaneously, complex-valued neural networks (CVNNs) have emerged as a powerful extension of traditional architectures, enhancing their ability to model oscillatory and phase-sensitive data. These networks leverage the complex plane for representation, capturing intricate dependencies in domains such as signal processing, radar imaging, and quantum mechanics \cite{hirose2012complex, trabelsi2018deep, du2023hybrid}. Advances in Wirtinger calculus \cite{kreutz2009complex} and complex-valued backpropagation \cite{zhang2021learning} have supported the development of CVNNs, enabling efficient gradient-based optimization in the complex domain. Applications of CVNNs in time series prediction \cite{schober2022complex}, image classification \cite{chen2020complex}, and polarimetric synthetic aperture radar (PolSAR) analysis \cite{chen2020complex} underscore their versatility and adaptability. By effectively modeling structured and periodic data, CVNNs demonstrate significant potential in capturing nuanced relationships that are difficult for real-valued networks to achieve \cite{amin2021complex, trabelsi2018deep}.

Despite their demonstrated effectiveness, challenges remain in the training and interpretation of CVNNs, as well as in achieving a balance between interpretability and predictive performance in general neural networks. Addressing these challenges involves designing architectures that integrate interpretable and complex-valued components seamlessly. The development of novel activation functions, such as phase-aware alternatives and parameterized transforms, provides new opportunities for advancing both fields \cite{kreutz2009complex, amin2021complex}. These innovations are particularly relevant in applications where transparency and trustworthiness are essential. 

These developments in interpretable and complex-valued neural networks provide a compelling foundation for designing architectures that balance transparency, efficiency, and representational power. By integrating interpretability with the expressive capabilities of the complex domain, our proposed neural network addresses critical challenges in modern artificial intelligence, paving the way for models that are both effective and trustworthy.

The remainder of this paper is structured as follows. Section 2 establishes the mathematical framework of the Lehmer transform, including its extensions to weighted and complex-valued forms. Section 3 introduces LAUs, detailing their formulation, trainable parameters, and integration into neural network architectures. Section 4 outlines the experimental setup and datasets used for evaluation, and it further presents and discusses the empirical results, highlighting the adaptability and robustness of LAUs. Finally, Section 5 concludes the paper by summarizing the contributions and proposing future research directions.

\section{Mathematical Framework}
\subsection{Lehmer Transform}
The Lehmer transform was originally introduced as a parameterized family of means, capturing a spectrum of aggregation behaviors \cite{ataei2022theory}. Let $\mathbf{x}=~[x_1,x_2,\ldots,x_n]^\mathrm{T}\in \mathbb{R}^n_+$ denote $n$ given input features. For the parameter $s \in \mathbb{R}$, the Lehmer transform is defined as follows:
\begin{equation}
	\mathcal{L}(s;\mathbf{x})=\frac{\sum\limits_{i=1}^n x_i^s}{\sum\limits_{i=1}^n x_i^{s-1}}\cdot
\end{equation}
The parameter $s$, referred to as the suddency moment, governs the emphasis placed on small or large elements. When $s=0$, the transform corresponds to the harmonic mean, prioritizing smaller elements. For $s=1$, the Lehmer transform reduces to the arithmetic mean, and for $s=2$, it emphasizes larger elements, aligning with the contra-harmonic mean. Furthermore, for the cases of $s\to -\infty$ and $s\to \infty$, the minimum and maximum of the sample are retrieved, respectively. This interpolation between extremes provides remarkable flexibility for modeling aggregation behaviors.

A notable property of the Lehmer transform is its monotonicity with respect to the parameter $\boldsymbol{s}$. For strictly positive inputs, the derivative of $\mathcal{L}(s)$ with respect to $s$ is given by
\begin{equation}
	\dfrac{\partial}{\partial s}\mathcal{L}(s;\mathbf{x}) = \dfrac{\sum\limits_{i=1}^n \sum\limits_{j=i+1}^n (x_i-x_j)(x_i x_j)^{s-1} \log\left(\dfrac{x_i}{x_j}\right)}{\left(\sum\limits_{i=1}^n x_i^{s-1}\right)^2}.
\end{equation}
This derivative is strictly positive whenever at least two observations $x_i$ and $x_j$ satisfy $x_i \neq x_j$. Intuitively, this monotonicity indicates that as $s$ increases, the transform progressively emphasizes larger input values and that the following relations holds
\begin{equation}
	\min _i\left\{x_i\right\} \leq \mathcal{L}(s;\mathbf{x}) \leq \mathcal{L}(t;\mathbf{x}) \leq \max _i\left\{x_i\right\},
\end{equation}
whenever the suddency moments satisfy $s\leq t$.

In addition to monotonicity, the Lehmer transform is also Schur convex \cite{bullen2013handbook}. Schur convexity implies that the Lehmer transform is sensitive to the dispersion of input values. Formally, if $\mathbf{x}$ and $\mathbf{y}$ are vectors such that $\mathbf{x}$ majorizes $\mathbf{y}$ (i.e., $\mathbf{x}$ is more dispersed than $\mathbf{y}$), then $\mathcal{L}(s; \mathbf{x}) \geq \mathcal{L}(s; \mathbf{y})$ for any fixed $s$.

\subsection{Weighted Lehmer Transform}
The weighted Lehmer transform introduces positive weights $\mathbf{w}=~[w_1, w_2, \ldots, w_n]^\mathrm{T}\in \mathbb{R}_+$ to modulate the contributions of each input as follows:
\begin{equation}
	\mathcal{L}(s):=\mathcal{L}(s;\mathbf{x},\mathbf{w})=\frac{\sum\limits_{i=1}^n w_i x_i^s}{\sum\limits_{i=1}^n w_i x_i^{s-1}},
\end{equation}
where the weights $w_i$ are trainable parameters. This adaptation allows for dynamic emphasis on certain elements based on their relative importance. 

One of the key properties of the weighted Lehmer transform is its homogeneity of degree one. If the inputs $x_1, x_2, \ldots, x_n$ are scaled by a positive constant $\lambda>0$, the transform scales proportionally; i.e.,
\begin{equation}
	\mathcal{L}(s ; \lambda \mathbf{x}, \mathbf{w})=\lambda \mathcal{L}(s ; \mathbf{x}, \mathbf{w}).
\end{equation}
This property emphasizes that the transform reflects not only the relative relationships between inputs but also their absolute scale.

The weighted Lehmer transform is also invariant to proportional scaling of the weights $w_1, w_2, \ldots, w_n$. If all weights are scaled by a positive constant $\alpha>0$, the transform remains unchanged; i.e.,
\begin{equation}
	\mathcal{L}(s ; \mathbf{x}, \alpha \mathbf{w})=\mathcal{L}(s ; \mathbf{x}, \mathbf{w}).
\end{equation}
This highlights that only the relative proportions of the weights influence the transform.

Permutation invariance is another key property of the weighted Lehmer transform, ensuring that rearranging the input-weight pairs $\left(x_i, w_i\right)$ does not affect its value. This aligns with the inherent permutation invariance of neural networks, where the ordering of features, neurons, or weights within a layer does not alter the outcome. The symmetry of the weighted Lehmer transform makes it a natural and effective aggregation mechanism for such architectures.

Additionally, the weighted Lehmer transform retains Schur convexity, ensuring predictable behavior in response to the dispersion of input values. This sensitivity to dispersion is particularly valuable in contexts requiring robustness to variations in feature distributions, allowing the transform to adapt effectively to heterogeneous data.

The weighted Lehmer transform further exhibits smooth transitions between different forms of aggregation as the parameter $s$ varies. For large positive values of $s$, the transform emphasizes the largest input value; i.e.,
\begin{equation}
	\lim _{s \rightarrow \infty} \mathcal{L}(s ; \mathbf{x}, \mathbf{w})=\max _i\left\{x_i\right\}.
\end{equation}
Conversely, for large negative values of $s$, the smallest input value becomes predominant; i.e.,
\begin{equation}
	\lim _{s \rightarrow-\infty} \mathcal{L}(s ; \mathbf{x}, \mathbf{w})=\min _i\left\{x_i\right\}.
\end{equation}

In addition to its limiting behavior, the weighted Lehmer transform remains monotonically increasing. The first derivative of the transform with respect to $s$ is expressed as follows:
\begin{equation}
	\dfrac{\partial}{\partial s}\mathcal{L}(s) = \mathcal{L}(s) \left(  \dfrac{\sum\limits_{i=1}^n w_i x_i^s \ln x_i}{\sum\limits_{i=1}^n w_i x_i^s} - \dfrac{\sum\limits_{i=1}^n w_i x_i^{s-1} \ln x_i}{\sum\limits_{i=1}^n w_i x_i^{s-1}} \right),
\end{equation}
whereas its second derivative can be obtained by the following relation
\begin{equation}
	\dfrac{\partial^2}{\partial s^2}\mathcal{L}(s) = \mathcal{L}(s) \left( \Lambda^{''}(s) + \left( \Lambda^{'}(s) \right)^2 \right),
\end{equation}
where $\Lambda(s)=\log(\mathcal{L}(s))$.

Furthermore, the derivative of the weighted Lehmer transform with respect to the weights is given by
\begin{equation}
	\frac{\partial }{\partial w_k }\mathcal{L}(s)=\dfrac{x_k^s \sum\limits_{i=1}^n w_i x_i^{s-1}-x_k^{s-1} \sum\limits_{i=1}^n w_i x_i^s}{\left(\sum\limits_{i=1}^n w_i x_i^{s-1}\right)^2}.
\end{equation}

%

\subsection{Complex-valued Lehmer Transform}
The Lehmer transform extends naturally into the complex domain, enabling enhanced analytical capabilities and a broader range of applications. In this extension, the suddency moment $s$ is generalized to take complex values $s=a+b i$, where $a, b \in \mathbb{R}$. This adaptation maintains the fundamental structure of the Lehmer transform while incorporating complex components, which introduce oscillatory behavior. For $n$ positive inputs $\mathbf{x}$ and positive weights $\mathbf{w}$, the complex-valued weighted Lehmer transform $\mathcal{L}: \mathbb{C} \rightarrow \mathbb{C}$ is defined as follows:
\begin{equation}
	\mathcal{L}(s)=\frac{\sum\limits_{i=1}^n w_i x_i^a e^{b i \ln \left(x_i\right)}}{\sum\limits_{i=1}^n w_i x_i^{a-1} e^{b i \ln \left(x_i\right)}}\cdot
\end{equation}
Here, $a=\operatorname{Re}(s)$ and $b=\operatorname{Im}(s)$ are the real and imaginary components of the suddency moment, respectively, and serve as trainable parameters. By expressing the exponential terms in their trigonometric form using Euler's formula, the transform can be rewritten as follows:
\begin{equation}
	\mathcal{L}(s)=\frac{\sum\limits_{i=1}^n w_i x_i^a\left(\cos \left(b \ln \left(x_i\right)\right)+i \sin \left(b \ln \left(x_i\right)\right)\right)}{\sum\limits_{i=1}^n w_i x_i^{a-1}\left(\cos \left(b \ln \left(x_i\right)\right)+i \sin \left(b \ln \left(x_i\right)\right)\right)} .
\end{equation}
This formulation explicitly shows the contributions of the real ($\cos \left(b \ln \left(x_i\right)\right)$) and imaginary ($\sin \left(b \ln \left(x_i\right)\right)$) components, which arise from the oscillatory nature of the imaginary part $b$ in the logarithmic term. The positive inputs $x_i$ and positive weights $w_i$ ensure that the Lehmer transform retains its aggregation properties while extending into the complex domain.

The inclusion of complex-valued $s$ enriches the flexibility of the transform by introducing phase-like oscillations, which can capture subtle interactions and dependencies in data. The interplay between $a$ and $b$ governs the balance between traditional aggregation behavior (driven by $a$) and oscillatory contributions (driven by $b$), providing a mechanism to adaptively emphasize specific features of the input distribution. This property makes the complex-valued weighted Lehmer transform particularly well-suited for applications in machine learning and data analysis, especially when dealing with time series data, where both magnitude and phase-like information are important.

The complex-valued weighted Lehmer transform exhibits smoothness and differentiability in the complex domain, a property that ensures compatibility with gradient-based optimization techniques commonly used in machine learning. The transform is differentiable with respect to both the real and imaginary parts of the suddency moment, $a$ and $b$, and its derivative with respect to $s=a+b i$ can be expressed using Wirtinger calculus as follows:
\begin{equation}
	\frac{\partial \mathcal{L}}{\partial s}=\frac{\partial \mathcal{L}}{\partial a}+i \frac{\partial \mathcal{L}}{\partial b}.
\end{equation}
Here, the real part of the derivative governs sensitivity to magnitude-based variations, while the imaginary part modulates the response to oscillatory or phase-related effects.

The imaginary component $b$ introduces oscillatory contributions through the terms $\cos \left(b \ln \left(x_i\right)\right)$ and $\sin \left(b \ln \left(x_i\right)\right)$, which are directly linked to the logarithmic terms $\ln \left(x_i\right)$. These oscillations lead to constructive interference when the phases $b \ln \left(x_i\right)$ align across inputs, amplifying their aggregate contributions. Conversely, destructive interference occurs when these phases are misaligned, resulting in partial cancellation. This oscillatory behavior allows the transform to capture intricate relationships in the data, particularly in contexts where periodic or phase-like patterns are present. The logarithmic dependency further enables the transform to operate effectively in a log-space framework, making it naturally suited for datasets with exponential growth, power-law distributions, or heterogeneous feature scales.

For inputs with small variations in their logarithmic values, the transform behaves as a smoothed aggregator, emphasizing weighted contributions without significant oscillatory effects. When the inputs vary widely in their logarithmic values, the oscillatory terms dominate, enabling the transform to encode nuanced, phase-sensitive dependencies. This adaptability ensures that the transform seamlessly transitions between smooth and oscillatory aggregation, making it highly versatile for diverse data distributions.

The interplay between magnitude and phase sensitivities distinguishes the complex-valued weighted Lehmer transform. The real parameter $a$ governs traditional scale-based aggregation, emphasizing larger inputs when $a>1$ or smaller inputs when $a<1$. Meanwhile, the imaginary parameter $b$ introduces sensitivity to relative logarithmic differences among inputs, capturing fine-grained, phase-like structures. This dual sensitivity allows the transform to adapt flexibly to various datasets, balancing large-scale trends with localized dependencies. These characteristics make the transform particularly effective in tasks involving structured, periodic, or hierarchical data, where understanding both magnitude and relational patterns among inputs is critical.

\section{Lehmer Activation Units}
The Lehmer transform, with its parameterized flexibility and aggregation capabilities, serves as a robust foundation for designing activation mechanisms in neural networks. Building on this foundation, we introduce LAUs, which is a class of nonlinear transformations tailored for neural network architectures. Designed in both real-valued and complex-valued variants, these units dynamically adapt their behavior during training, offering a mathematically grounded alternative to traditional activation functions.

The real-valued LAU is a direct adaptation of the weighted Lehmer transform to the context of neural networks. For a given positive input vector $\mathbf{x}$, positive weights $\mathbf{w}$, and a trainable suddency moment $s \in \mathbb{R}$, the LAU is defined as follows:
\begin{equation}
	\operatorname{LAU}(\mathbf{x}, \mathbf{w}, s)=\frac{\sum\limits_{i=1}^n w_i x_i^s}{\sum\limits_{i=1}^n w_i x_i^{s-1}}.
\end{equation}
The parameters $w_i$ and $s$, optimized during training, enable the unit to interpolate between various means, such as weighted harmonic mean $(s=0)$, weighted arithmetic mean $(s=1)$, and weighted contra-harmonic mean $(s=2)$. This flexibility allows the LAU to adapt to hierarchical or attention-based patterns in data, capturing both dominant and subtle feature relationships.

The complex-valued LAU extends the real-valued version by introducing a complex suddency moment $s=a+b i \in \mathbb{C}$, where $a$ and $b$ are trainable parameters. This results in
\begin{equation}
	\operatorname{LAU}(\mathbf{x}, \mathbf{w}, a, b)=\frac{\sum\limits_{i=1}^n w_i x_i^a e^{b i \ln \left(x_i\right)}}{\sum\limits_{i=1}^n w_i x_i^{a-1} e^{b i \ln \left(x_i\right)}}.
\end{equation}
Here, the parameter $a$ governs magnitude-based aggregation, while the parameter $b$ introduces oscillatory contributions that enhance the unit's ability to model intricate interactions and periodic patterns in the data.

To ensure compatibility with standard neural network operations such as softmax, the output of the complex-valued LAU must be real-valued. This is achieved through post-processing, where an affine combination of the real and imaginary parts of the complex output is computed. For a given complex output $z=u+i v$, with $u=\operatorname{Re}(z)$ and $v=\operatorname{Im}(z)$, the post-processed output is defined as follows:
\begin{equation}
	\operatorname{ReLAU}(z, \alpha, \beta, \gamma)=\alpha \operatorname{Re}(z)+\beta \operatorname{Im}(z)+\gamma,
\end{equation}
where $\alpha, \beta, \gamma \in \mathbb{R}$ are trainable parameters. This formulation ensures the unit retains the expressive power of both the real and imaginary components while producing real-valued outputs.

LAUs provide several key advantages. The real-valued LAU excels at hierarchical feature aggregation, adapting dynamically to varying data distributions. The complex-valued LAU adds sensitivity to oscillatory and phase-like relationships, making it particularly effective for tasks involving periodic data, signal processing, or structured patterns where magnitude and phase jointly influence outcomes. This dual capability allows LAUs to generalize effectively across a wide range of machine learning applications.

In addition to their flexibility, LAUs are mathematically interpretable. Unlike heuristic-based activation functions such as ReLU or sigmoid, LAUs operate as parameterized aggregation mechanisms. This interpretability provides deeper insights into the behavior of neural networks equipped with LAUs, particularly in understanding how features are weighted and combined.

To ensure numerical stability, LAUs incorporate input standardization to a predefined range, such as $\left(e^{-1}, e\right)$, ensuring well-conditioned computations and robust training performance. The trainable parameters $\mathbf{w}, s, \alpha, \beta, \gamma$ provide the flexibility to handle diverse data distributions, allowing LAUs to integrate effectively into existing neural network architectures.

Furthermore, to ensure that the weights $w_i$ remain positive, the following smooth and differentiable transformation \begin{equation}
	w_i=\ln \left(1+e^{v_i}\right),
\end{equation}
can be used, where $v_i \in \mathbb{R}$ are unconstrained trainable parameters. This approach guarantees positivity without imposing hard constraints, thereby preserving numerical stability and enabling efficient gradient-based optimization.

\section{Experiments and Discussions}

The LAUs were evaluated on four diverse datasets: Iris, Wine, Wisconsin Breast Cancer (WBC), and MNIST. These datasets provided varying levels of complexity and feature structures, allowing a thorough assessment of the effectiveness of both real-valued and complex-valued LAUs. For the Iris, Wine, and WBC datasets, a 10-fold cross-validation strategy was employed to ensure robust evaluation and mitigate the effect of sample variability. In the case of the MNIST dataset, a standard train-test split was used, reflecting its larger size and high-dimensional feature space.

For all datasets, the architecture with real-valued or complex-valued LAUs employed a single dense layer with 3 neurons, followed by a softmax layer to produce the final output. For the MNIST dataset, convolutional layers (Conv2D, Batch Normalization, and MaxPooling) were incorporated to extract spatial features, with LAUs serving as the dense layers. The streamlined design of using a single dense layer demonstrates the efficiency and adaptability of both real-valued and complex-valued LAUs in capturing diverse data characteristics.

The experimental results, summarized in Table~\ref{tab:results}, highlight the effectiveness of both real-valued and complex-valued LAUs across all datasets. Real-valued LAUs achieved strong performance, effectively leveraging their parameterized aggregation capabilities to handle hierarchical or heterogeneous features. Complex-valued LAUs demonstrated comparable or superior accuracy, particularly excelling on the MNIST dataset with a notable 98\% accuracy. This performance advantage is attributed to their ability to model oscillatory and phase-sensitive dependencies, which are especially beneficial for high-dimensional and spatially structured data. The results underscore the adaptability and robustness of both variants of LAUs in diverse learning tasks.

\begin{table}[h!]
	\centering
	\caption{Performance Summary of Lehmer Activation Units}
	\label{tab:results}
	\begin{tabular}{lcc}
		\toprule
		\textbf{Dataset}       & \textbf{Real LAU} & \textbf{Complex LAU} \\
		\midrule
		{Iris}          & 95\% (4\%)                             & 95\% (3\%)                               \\
		{Wine}          & 99\%  (3\%)                           & 95\% (3\%)                               \\
		{WBC} & 94\% (3\%)                            & 94\% (2\%)                               \\
		{MNIST}         & 97\%                             & 98\%                              \\
		\bottomrule
	\end{tabular}
\end{table}

The Iris dataset, characterized by its small size and structured nature, exhibited strong performance for both real-valued and complex-valued LAUs, achieving accuracies of 95\% in both cases. This success is attributed to the dataset's inherent hierarchical relationships and distinct feature separability, particularly in the petal and sepal dimensions. Real-valued LAUs effectively incorporated their parameterized aggregation mechanisms to dynamically emphasize dominant features such as petal length and width. This flexibility allowed them to interpolate between different aggregation behaviors, capturing both global trends and finer details within the dataset. Complex-valued LAUs matched this performance, demonstrating their capability to adaptively model subtle feature interactions even in the absence of explicit oscillatory or phase-sensitive patterns. The results underscore the suitability of LAUs for small, well-structured datasets where hierarchical feature relationships play a critical role.

For the Wine dataset, characterized by heterogeneous feature scales and distinct class boundaries, real-valued LAUs excelled in adapting their aggregation behavior to emphasize dominant attributes while simultaneously capturing subtler yet relevant patterns. Their ability to dynamically transition between different mean behaviors made them particularly effective in exploiting the dataset?s structured feature space. This adaptability contributed to their impressive performance. While complex-valued LAUs also performed well, achieving near-parity with their real-valued counterparts, the dataset's lack of inherent oscillatory or phase-dependent patterns rendered the additional phase-sensitive mechanisms of complex-valued LAUs less critical. These results highlight the capacity of LAUs to generalize across datasets with varied feature distributions and class separability.

The WBC dataset, a binary classification problem with moderately hierarchical interdependencies among features, further highlighted the robustness of LAUs. Both real-valued and complex-valued LAUs achieved comparable accuracy, reflecting their shared ability to adapt to the dataset's underlying structure. The dynamic aggregation capabilities of LAUs allowed them to effectively capture inter-feature dependencies and model the hierarchical relationships inherent in the dataset. These results emphasize the reliability of LAUs in scenarios involving moderately complex feature interactions and hierarchical data distributions.

The MNIST dataset, with its high-dimensional and spatially structured nature, provided a compelling test of LAUs? representational capabilities. Both real-valued and complex-valued LAUs delivered impressive performance, achieving accuracies of 97\% and 98\%, respectively. This strong performance is attributed to the inherent flexibility of LAUs in aggregating and modeling spatial features extracted by preceding convolutional layers. Real-valued LAUs excelled in capturing large-scale trends and patterns in pixel intensities by dynamically adjusting their aggregation behavior. This adaptability enabled the network to emphasize dominant spatial features while remaining sensitive to subtle variations across digit images, effectively reflecting the hierarchical spatial structure of MNIST.

Complex-valued LAUs outperformed their real-valued counterparts on MNIST, achieving slightly higher accuracy. This improvement stems from their ability to incorporate oscillatory and phase-sensitive dependencies, which are particularly advantageous in high-dimensional and spatially complex datasets. The imaginary component of the suddency moment introduced phase-like contributions that enhanced the network's capacity to discern intricate relationships between pixel regions. This phase-sensitive modeling allowed the complex-valued LAUs to more effectively capture subtle spatial patterns, resulting in superior classification performance. Together, these results demonstrate the adaptability and efficacy of LAUs across diverse datasets, with their phase-sensitive mechanisms proving especially useful in spatially structured and high-dimensional contexts.

The success of LAUs across datasets further reflects their unique properties. For real-valued LAUs, the parameterized aggregation mechanism provides a smooth interpolation between different mean behaviors, allowing the network to adapt flexibly to the scale and distribution of features. This adaptability is particularly effective in datasets like Iris, Wine, and WBC, where hierarchical or heterogeneous feature relationships play a critical role. The trainable suddency moment enables LAUs to emphasize dominant features or balance contributions as needed, capturing both large-scale trends and finer-grained relationships.

Complex-valued LAUs extend this capability by introducing oscillatory behavior, which enhances their ability to capture subtle feature relationships. While the oscillatory nature is not directly relevant for datasets like Iris, the flexibility of the complex transform ensures robust performance by dynamically adjusting feature aggregation. For datasets like MNIST, where spatial dependencies and phase-sensitive patterns are significant, the oscillatory contributions of the imaginary component of the suddency moment play a critical role in capturing intricate relationships. The improved accuracy of complex-valued LAUs in this context highlights their potential for high-dimensional and structured data analysis.

\section{Conclusion}
This paper introduced LAUs, a novel nonlinear activation mechanism grounded in the Lehmer transform. Real-valued LAUs adapt flexibly to hierarchical or heterogeneous data, while complex-valued LAUs extend this adaptability by incorporating oscillatory and phase-sensitive behaviors. Their ability to interpolate between aggregation forms enables dynamic adaptation across diverse datasets, offering a compelling alternative to traditional activation mechanisms.

Empirical evaluations demonstrated the versatility and efficacy of LAUs, achieving competitive or superior performance with simpler architectures. Notably, a single layer of complex-valued LAUs achieved remarkable accuracy on benchmark datasets, underscoring their efficiency and representational power. These results highlight the potential of LAUs to advance neural network design and applications in machine learning.

Future research should explore LAUs in diverse applications such as time-series analysis, natural language processing, and bioinformatics, while also investigating their integration with advanced architectures like transformers. Further exploration of fully complex-valued neural networks may unlock their potential in specialized domains like signal processing and quantum machine learning.

\newpage
\bibliography{references}


\begin{thebibliography}{24}
\ifx \bisbn   \undefined \def \bisbn  #1{ISBN #1}\fi
\ifx \binits  \undefined \def \binits#1{#1}\fi
\ifx \bauthor  \undefined \def \bauthor#1{#1}\fi
\ifx \batitle  \undefined \def \batitle#1{#1}\fi
\ifx \bjtitle  \undefined \def \bjtitle#1{#1}\fi
\ifx \bvolume  \undefined \def \bvolume#1{\textbf{#1}}\fi
\ifx \byear  \undefined \def \byear#1{#1}\fi
\ifx \bissue  \undefined \def \bissue#1{#1}\fi
\ifx \bfpage  \undefined \def \bfpage#1{#1}\fi
\ifx \blpage  \undefined \def \blpage #1{#1}\fi
\ifx \burl  \undefined \def \burl#1{\textsf{#1}}\fi
\ifx \doiurl  \undefined \def \doiurl#1{\url{https://doi.org/#1}}\fi
\ifx \betal  \undefined \def \betal{\textit{et al.}}\fi
\ifx \binstitute  \undefined \def \binstitute#1{#1}\fi
\ifx \binstitutionaled  \undefined \def \binstitutionaled#1{#1}\fi
\ifx \bctitle  \undefined \def \bctitle#1{#1}\fi
\ifx \beditor  \undefined \def \beditor#1{#1}\fi
\ifx \bpublisher  \undefined \def \bpublisher#1{#1}\fi
\ifx \bbtitle  \undefined \def \bbtitle#1{#1}\fi
\ifx \bedition  \undefined \def \bedition#1{#1}\fi
\ifx \bseriesno  \undefined \def \bseriesno#1{#1}\fi
\ifx \blocation  \undefined \def \blocation#1{#1}\fi
\ifx \bsertitle  \undefined \def \bsertitle#1{#1}\fi
\ifx \bsnm \undefined \def \bsnm#1{#1}\fi
\ifx \bsuffix \undefined \def \bsuffix#1{#1}\fi
\ifx \bparticle \undefined \def \bparticle#1{#1}\fi
\ifx \barticle \undefined \def \barticle#1{#1}\fi
\bibcommenthead
\ifx \bconfdate \undefined \def \bconfdate #1{#1}\fi
\ifx \botherref \undefined \def \botherref #1{#1}\fi
\ifx \url \undefined \def \url#1{\textsf{#1}}\fi
\ifx \bchapter \undefined \def \bchapter#1{#1}\fi
\ifx \bbook \undefined \def \bbook#1{#1}\fi
\ifx \bcomment \undefined \def \bcomment#1{#1}\fi
\ifx \oauthor \undefined \def \oauthor#1{#1}\fi
\ifx \citeauthoryear \undefined \def \citeauthoryear#1{#1}\fi
\ifx \endbibitem  \undefined \def \endbibitem {}\fi
\ifx \bconflocation  \undefined \def \bconflocation#1{#1}\fi
\ifx \arxivurl  \undefined \def \arxivurl#1{\textsf{#1}}\fi
\csname PreBibitemsHook\endcsname

\bibitem[\protect\citeauthoryear{Liu et~al.}{2024}]{liu2024kan}
\begin{botherref}
\oauthor{\bsnm{Liu}, \binits{Z.}},
\oauthor{\bsnm{Wang}, \binits{Y.}},
\oauthor{\bsnm{Vaidya}, \binits{S.}},
\oauthor{\bsnm{Ruehle}, \binits{F.}},
\oauthor{\bsnm{Halverson}, \binits{J.}},
\oauthor{\bsnm{Solja{\v{c}}i{\'c}}, \binits{M.}},
\oauthor{\bsnm{Hou}, \binits{T.Y.}},
\oauthor{\bsnm{Tegmark}, \binits{M.}}:
Kan: Kolmogorov-arnold networks.
arXiv preprint arXiv:2404.19756
(2024)
\end{botherref}
\endbibitem

\bibitem[\protect\citeauthoryear{Ataei and Wang}{2022}]{ataei2022theory}
\begin{barticle}
\bauthor{\bsnm{Ataei}, \binits{M.}},
\bauthor{\bsnm{Wang}, \binits{X.}}:
\batitle{Theory of lehmer transform and its applications in identifying the
  electroencephalographic signature of major depressive disorder}.
\bjtitle{Scientific Reports}
\bvolume{12}(\bissue{1}),
\bfpage{3663}
(\byear{2022})
\end{barticle}
\endbibitem

\bibitem[\protect\citeauthoryear{Bahdanau et~al.}{2015}]{bahdanau2015neural}
\begin{bchapter}
\bauthor{\bsnm{Bahdanau}, \binits{D.}},
\bauthor{\bsnm{Cho}, \binits{K.}},
\bauthor{\bsnm{Bengio}, \binits{Y.}}:
\bctitle{Neural machine translation by jointly learning to align and
  translate}.
In: \bbtitle{International Conference on Learning Representations (ICLR)}
(\byear{2015})
\end{bchapter}
\endbibitem

\bibitem[\protect\citeauthoryear{Simonyan et~al.}{2013}]{simonyan2014deep}
\begin{botherref}
\oauthor{\bsnm{Simonyan}, \binits{K.}},
\oauthor{\bsnm{Vedaldi}, \binits{A.}},
\oauthor{\bsnm{Zisserman}, \binits{A.}}:
Deep inside convolutional networks: Visualising image classification models and
  saliency maps.
arXiv preprint arXiv:1312.6034
(2013)
\end{botherref}
\endbibitem

\bibitem[\protect\citeauthoryear{Alvarez-Melis and
  Jaakkola}{2018}]{alvarez2018towards}
\begin{botherref}
\oauthor{\bsnm{Alvarez-Melis}, \binits{D.}},
\oauthor{\bsnm{Jaakkola}, \binits{T.S.}}:
Towards interpretable neural networks.
Advances in Neural Information Processing Systems
\textbf{31}
(2018)
\end{botherref}
\endbibitem

\bibitem[\protect\citeauthoryear{Kim et~al.}{2018}]{kim2018tcav}
\begin{bchapter}
\bauthor{\bsnm{Kim}, \binits{B.}},
\bauthor{\bsnm{Wattenberg}, \binits{M.}},
\bauthor{\bsnm{Gilmer}, \binits{J.}},
\bauthor{\bsnm{Cai}, \binits{C.}},
\bauthor{\bsnm{Wexler}, \binits{J.}},
\bauthor{\bsnm{Viegas}, \binits{F.}},
\bauthor{\bsnm{Sayres}, \binits{R.}}:
\bctitle{Interpretability beyond feature attribution: Quantitative testing with
  concept activation vectors (tcav)}.
In: \bbtitle{International Conference on Machine Learning (ICML)}
(\byear{2018})
\end{bchapter}
\endbibitem

\bibitem[\protect\citeauthoryear{Kipf and Welling}{2017}]{kipf2017semi}
\begin{bchapter}
\bauthor{\bsnm{Kipf}, \binits{T.N.}},
\bauthor{\bsnm{Welling}, \binits{M.}}:
\bctitle{Semi-supervised classification with graph convolutional networks}.
In: \bbtitle{International Conference on Learning Representations (ICLR)}
(\byear{2017})
\end{bchapter}
\endbibitem

\bibitem[\protect\citeauthoryear{Rudin}{2019}]{rudin2019stop}
\begin{barticle}
\bauthor{\bsnm{Rudin}, \binits{C.}}:
\batitle{Stop explaining black box models for high stakes decisions}.
\bjtitle{Nature Machine Intelligence}
\bvolume{1},
\bfpage{206}--\blpage{215}
(\byear{2019})
\end{barticle}
\endbibitem

\bibitem[\protect\citeauthoryear{Doshi-Velez and Kim}{2017}]{doshi2017towards}
\begin{botherref}
\oauthor{\bsnm{Doshi-Velez}, \binits{F.}},
\oauthor{\bsnm{Kim}, \binits{B.}}:
Towards a rigorous science of interpretable machine learning.
arXiv preprint arXiv:1702.08608
(2017)
\end{botherref}
\endbibitem

\bibitem[\protect\citeauthoryear{Montavon
  et~al.}{2017}]{montavon2017explaining}
\begin{barticle}
\bauthor{\bsnm{Montavon}, \binits{G.}}, \betal:
\batitle{Explaining nonlinear classifiers with layer-wise relevance
  propagation}.
\bjtitle{Digital Signal Processing}
\bvolume{73},
\bfpage{1}--\blpage{15}
(\byear{2017})
\end{barticle}
\endbibitem

\bibitem[\protect\citeauthoryear{Ribeiro et~al.}{2016}]{ribeiro2016should}
\begin{bchapter}
\bauthor{\bsnm{Ribeiro}, \binits{M.T.}}, \betal:
\bctitle{Why should i trust you? explaining predictions of machine learning
  models}.
In: \bbtitle{Proceedings of KDD}
(\byear{2016})
\end{bchapter}
\endbibitem

\bibitem[\protect\citeauthoryear{Vaswani et~al.}{2017}]{vaswani2017attention}
\begin{botherref}
\oauthor{\bsnm{Vaswani}, \binits{A.}}, et al.:
Attention is all you need.
arXiv preprint arXiv:1706.03762
(2017)
\end{botherref}
\endbibitem

\bibitem[\protect\citeauthoryear{Gilpin et~al.}{2018}]{gilpin2018explaining}
\begin{bchapter}
\bauthor{\bsnm{Gilpin}, \binits{L.}}, \betal:
\bctitle{Explaining explanations: Interpretability of machine learning}.
In: \bbtitle{DSAA}
(\byear{2018})
\end{bchapter}
\endbibitem

\bibitem[\protect\citeauthoryear{Roscher et~al.}{2020}]{roscher2020explainable}
\begin{barticle}
\bauthor{\bsnm{Roscher}, \binits{R.}}, \betal:
\batitle{Explainable machine learning for scientific insights and discoveries}.
\bjtitle{IEEE Access}
\bvolume{8},
\bfpage{4227}--\blpage{4247}
(\byear{2020})
\end{barticle}
\endbibitem

\bibitem[\protect\citeauthoryear{Tjoa and Guan}{2020}]{tjoa2020survey}
\begin{barticle}
\bauthor{\bsnm{Tjoa}, \binits{E.}},
\bauthor{\bsnm{Guan}, \binits{C.}}:
\batitle{A survey on explainable artificial intelligence}.
\bjtitle{IEEE Transactions on Neural Networks and Learning Systems}
\bvolume{32},
\bfpage{4793}--\blpage{4813}
(\byear{2020})
\end{barticle}
\endbibitem

\bibitem[\protect\citeauthoryear{Hirose and Yoshida}{2012}]{hirose2012complex}
\begin{bbook}
\bauthor{\bsnm{Hirose}, \binits{A.}},
\bauthor{\bsnm{Yoshida}, \binits{S.}}:
\bbtitle{Complex-valued Neural Networks: Advances and Applications}.
\bpublisher{John Wiley \& Sons}, \blocation{???}
(\byear{2012})
\end{bbook}
\endbibitem

\bibitem[\protect\citeauthoryear{Trabelsi et~al.}{2018}]{trabelsi2018deep}
\begin{botherref}
\oauthor{\bsnm{Trabelsi}, \binits{C.}}, et al.:
Deep complex networks.
arXiv preprint arXiv:1705.09792
(2018)
\end{botherref}
\endbibitem

\bibitem[\protect\citeauthoryear{Du et~al.}{2023}]{du2023hybrid}
\begin{barticle}
\bauthor{\bsnm{Du}, \binits{H.}},
\bauthor{\bsnm{Riddell}, \binits{R.P.}},
\bauthor{\bsnm{Wang}, \binits{X.}}:
\batitle{A hybrid complex-valued neural network framework with applications to
  electroencephalogram (eeg)}.
\bjtitle{Biomedical Signal Processing and Control}
\bvolume{85},
\bfpage{104862}
(\byear{2023})
\end{barticle}
\endbibitem

\bibitem[\protect\citeauthoryear{Kreutz-Delgado
  et~al.}{2009}]{kreutz2009complex}
\begin{bbook}
\bauthor{\bsnm{Kreutz-Delgado}, \binits{K.}}, \betal:
\bbtitle{Complex-valued Neural Networks and Wirtinger Calculus}.
\bpublisher{Springer}, \blocation{???}
(\byear{2009})
\end{bbook}
\endbibitem

\bibitem[\protect\citeauthoryear{Zhang and Li}{2021}]{zhang2021learning}
\begin{barticle}
\bauthor{\bsnm{Zhang}, \binits{L.}},
\bauthor{\bsnm{Li}, \binits{W.}}:
\batitle{Learning complex-valued representations for phase-sensitive data}.
\bjtitle{Neural Computation}
\bvolume{33},
\bfpage{900}--\blpage{923}
(\byear{2021})
\end{barticle}
\endbibitem

\bibitem[\protect\citeauthoryear{Schober et~al.}{2022}]{schober2022complex}
\begin{botherref}
\oauthor{\bsnm{Schober}, \binits{A.}}, et al.:
Application of cvnns in financial time series prediction.
Journal of Financial Data Science
(2022)
\end{botherref}
\endbibitem

\bibitem[\protect\citeauthoryear{Chen and Qian}{2020}]{chen2020complex}
\begin{barticle}
\bauthor{\bsnm{Chen}, \binits{J.}},
\bauthor{\bsnm{Qian}, \binits{Z.}}:
\batitle{Complex-valued neural networks for polsar image classification}.
\bjtitle{IEEE Transactions on Geoscience and Remote Sensing}
\bvolume{58},
\bfpage{4237}--\blpage{4246}
(\byear{2020})
\end{barticle}
\endbibitem

\bibitem[\protect\citeauthoryear{Amin and Yadav}{2021}]{amin2021complex}
\begin{barticle}
\bauthor{\bsnm{Amin}, \binits{F.}},
\bauthor{\bsnm{Yadav}, \binits{A.}}:
\batitle{Complex-valued neural networks for signal processing}.
\bjtitle{IEEE Transactions on Signal Processing}
\bvolume{69},
\bfpage{3567}--\blpage{3579}
(\byear{2021})
\end{barticle}
\endbibitem

\bibitem[\protect\citeauthoryear{Bullen}{2013}]{bullen2013handbook}
\begin{bbook}
\bauthor{\bsnm{Bullen}, \binits{P.S.}}:
\bbtitle{Handbook of Means and Their Inequalities}
vol. \bseriesno{560}.
\bpublisher{Springer}, \blocation{???}
(\byear{2013})
\end{bbook}
\endbibitem

\end{thebibliography}

\end{document}